\documentclass[runningheads]{llncs}
\usepackage{graphicx}
\usepackage{amsmath}
\usepackage{amsfonts}
\usepackage{cite}
\usepackage[margin=1.5in]{geometry}

\usepackage{hyperref}

\def\eg{\emph{e.g.}}

\def\etal{\emph{et al.}}

\begin{document}
	
\title{Automatic 4D Facial Expression Recognition via Collaborative Cross-domain Dynamic Image Network}
\titlerunning{Automatic 4D Facial Expression Recognition via CCDN}
%
\author{Muzammil Behzad \and
Nhat Vo\and
Xiaobai Li \and
Guoying Zhao}

\authorrunning{\href{http://muzammilbehzad.com/}{M. Behzad \etal}}
%
\institute{Center for Machine Vision and Signal Analysis (CMVS), University of Oulu, Finland\\
Email: \{muzammil.behzad, nhat.vo, xiaobai.li, guoying.zhao\}@oulu.fi}

\maketitle              
\setcounter{footnote}{0} 
\begin{abstract}
	This paper proposes a novel 4D Facial Expression Recognition (FER) method using Collaborative Cross-domain Dynamic Image Network (CCDN). Given a 4D data of face scans, we first compute its geometrical images, and then combine their correlated information in the proposed cross-domain image representations. The acquired set is then used to generate cross-domain dynamic images (CDI) via rank pooling that encapsulates facial deformations over time in terms of a single image. For the training phase, these CDIs are fed into an end-to-end deep learning model, and the resultant predictions collaborate over multi-views for performance gain in expression classification. Furthermore, we propose a 4D augmentation scheme that not only expands the training data scale but also introduces significant facial muscle movement patterns to improve the FER performance. Results from extensive experiments on the commonly used BU-4DFE dataset under widely adopted settings show that our proposed method outperforms the state-of-the-art 4D FER methods by achieving an accuracy of $96.5\%$ indicating its effectiveness.
\end{abstract}

\section{Introduction}
\label{sec:intro}
Facial expressions (FEs) are one of the affective cues that play a vital role for humans to communicate socially via their emotions on daily basis. Driven by this, and the fact that there has been a surge in the research and development of computer vision based human-machine interactions, the artificially intelligent devices aim to better analyze, understand, learn and then mimic such facial expressions. Consequently, the research on facial expression recognition (FER) has attracted tremendous acknowledgment due to its significance in potential application fields like security, psychology, computing technology, bio-medical and education. In this regard, however, the pioneer contribution dates back to 1970s, when Ekman and Friesen~\cite{ekman1971constants} presented and analyzed the six universal human facial expressions: anger, disgust, fear, happiness, sadness, and surprise.

Over the past couple of decades, the FER community witnessed rise of many machine learning algorithms based on 2D static and dynamic face images (or video of 2D images). However, due to the diversity, complexity and subtlety of facial expressions, automatic FER remains a challenging task \cite{fasel2003automatic}. One possible reason is that 2D images are sensitive to pose variations and illumination conditions, driving the current 2D images based methods unstable. To combat this, 3D FER has motivated various new research directions with the help of emerging high-resolution and high-speed 3D imagery apparatus. This is mainly because 3D face data brings about a handsome amount of information  which constitutes essential cues to figure out facial movements over the facial depth-axis. However, this equally impacts the learning models at the cost of extremely challenging complexity. 

Naturally, since every facial expression is a resultant of various facial muscle movements, ultimately causing facial deformations \cite{5771466}, this encapsulated information is better caught in the geometric domain \cite{7457243}. As a result, the 3D face scans are really handy in decoding such movements, and hence, predicting the facial expressions. This is supported by the release of various complex and large-size facial databases containing multiple terabytes of such 3D static face scans. However, the delivery of famous BU-3DFE \cite{yin20063d} and Bosphorus~\cite{savran2008bosphorus} databases has paved a pioneered way to investigate FER using such 3D static face scans. An extension towards sequence of 3D face scans (referred as 3D videos or 4D) has ignited a much more interesting direction since it incorporates both the depth information in the facial geometry as well as its spatial movements across the temporal domain. The release of popular BU-4DFE database \cite{4813324} has allowed for swift developments in the study of 4D FER to precisely analyze the facial expressions using extended spatio-temporal information.

\subsection{Related Work}
As opposed to 3D FER, which does not contain the temporal information over geometrical domains \cite{5597896,6460694,zhen2015muscular,li2015efficient}, 4D facial data allows to capture an in-depth knowledge about the facial deformation patterns encoding a specific facial expression. Sun \etal~\cite{Sun:2010:TVF:1820799.1820803} worked out a way to generate correspondence between the 3D face scans over time. Driven by these correspondences, they proposed the idea of using spatio-temporal Hidden Markov Models (HMM) that capture the facial muscle movements by analyzing both inter-frame and intra-frame variations. Similarly, Yin \etal~\cite{4813324} utilized a 2D HMM to learn the facial deformations in the temporal domain for expression classification.

In another attempt based mainly on Riemannian analysis, Drira and Amor \cite{6460329, amor20144} introduced a deformation vector field combined with random forests that learns the local face deformation patterns along time. Similarly, Sandbach \etal~\cite{5771434,sandbach2012recognition} expressed information between neighboring 3D frames as motion-based features, known as Free-Form Deformation (FFD). Subsequently, they adopted the HMM and GentleBoost classifiers for FER. 

To classify FEs using Support Vector Machine (SVM) with a Radial Basis Function kernel, Fang \etal~\cite{FANG2012738} extracted two types of feature vectors represented as geometrical coordinates and its normal. In another of their work \cite{6130440}, they first exploited MeshHOG to calibrate the given face meshes. Afterwards, the dynamic Local Binary Patterns (LBP) were applied to capture deformations over time, followed by SVM for FER. Likewise, the authors of \cite{6553746} proposed a spatio-temporal feature that uses LBP to extract information encapsulated in the histogram of different facial regions as polar angles and curvatures.

Yao \etal~\cite{Yao:2018:TGS:3190503.3131345} applied the scattering operator \cite{bruna2013invariant} over 4D data, producing geometrical and textual scattering representations. Multiple Kernel Learning (MKL) was then applied to learn from this information for FER. Authors in \cite{fabiano2018spontaneous} presented a statistical shape model with global and local constraints in an attempt to recognize FEs. They claimed that the combination of global face shape and local shape index-based information can be handy for FER. Li \etal~\cite{8373807} introduced a Dynamic Geometrical Image Network (DGIN) for automatically recognizing expressions. Given 4D data, the differential geometry quantities are estimated first, followed by generating geometrical images. These images are then fed into the DGIN for an end-to-end training. The prediction results are based on fusing the predicted scores of different types of geometrical images.

\subsection{Motivation and Contribution}
The limitations incurred by 2D FER methods have been largely diminished by using the 3D face scans since they provide sufficient information about facial muscle movements. However, very handful 3D based methods have been extended towards 4D FER with some promising results. Since 4D data contains extremely significant patterns both in shape and temporal domains, these dynamic facial deformations play a pivotal role in FER. Contrary to the existing methods (\eg, \cite{8373807}), we believe that instead of independently tuning different parts of a given FER solution, appropriate representations of facial patterns, in both spatial and temporal domains, are crucial for learning correct network weights.

Motivated by this, in this paper, we aim to extract the facial deformation patterns jointly across multi-views and different geometrical domains, and propose a collaborative end-to-end deep network for 4D FER, namely Collaborative Cross-domain Dynamic Image Network (CCDN). We first extract texture and depth images for every given 3D face scan, and then propose a cross-domain (CD) representation where the correlation among different geometrical images are combined into a single image. Afterwards, the sequence of these cross-domain images are then used to compute dynamic images \cite{bilen2018action} for encapsulating the facial muscle movements over time. These cross-domain dynamic images (CDI) then collaborate over multi-views to improve FER by magnifying the correct probabilities. To the best of our knowledge, this is the first multi-view deep learning framework for 4D FER.

As well, it is a globally-accepted fact that the success of deep models rely considerably on the availability of large-scale data in order to learn patterns efficiently. The amount of data available for 4D FER is extremely insufficient, unfortunately. The commonly used BU-4DFE \cite{4813324} database for 4D FER contains 101 video samples for each of the six prototypical expression (total 606 3D face videos). Consequently, we propose a novel training-free 4D data augmentation scheme to expand both the training size and patterns. We compute the motion magnified versions of the original video samples and its reverse-ordered samples over multi-view data. These clips are then further duplicated into its flipped and rotated versions. A follow-up windowing method helps both convolutional and fully connected layers to learn extensively from the dynamic information.

Finally, we carry out extensive experiments on the BU-4DFE benchmark to showcase the performance gain of our method in recognizing facial expressions via 4D facial data. The rest of the paper is organized as follows: Section \ref{sec:proposedmethod} explains the working flow of the proposed CCDN method for automatic 4D FER including the generation of cross-domain dynamic images and the multi-views collaboration for maximizing classification performance. In Section \ref{sec:results}, we report the results of our extensive experiments both for 4D FER as well as 4D augmentation. Finally, Section \ref{sec:conclusions} concludes the paper.

\section{Proposed Automatic 4D FER Method}
\label{sec:proposedmethod}
In this section, we explain the working flow of our proposed automatic 4D FER method. First, we discuss the pre-processing steps to filter out unwanted components, and then obtain various geometrical images across multi-views. Second, we introduce the cross-domain and its dynamic images. Third, the collaborative scheme for 4D FER is elaborated. At last, we present our novel training-free augmentation method for 4D facial data. An overview of our proposed method for 4D FER is presented in Fig. \ref{fig:4DFER_floechart}.

\subsection{Pre-processing and Multi-views}
Raw data from BU-4DFE are noisy and contain interference components such as head hair, noisy vertices and regions apart from the face, which may cause trouble during deep model training, and hence, need to be filtered out.

\begin{figure}[t!]
	\includegraphics[width=\linewidth]{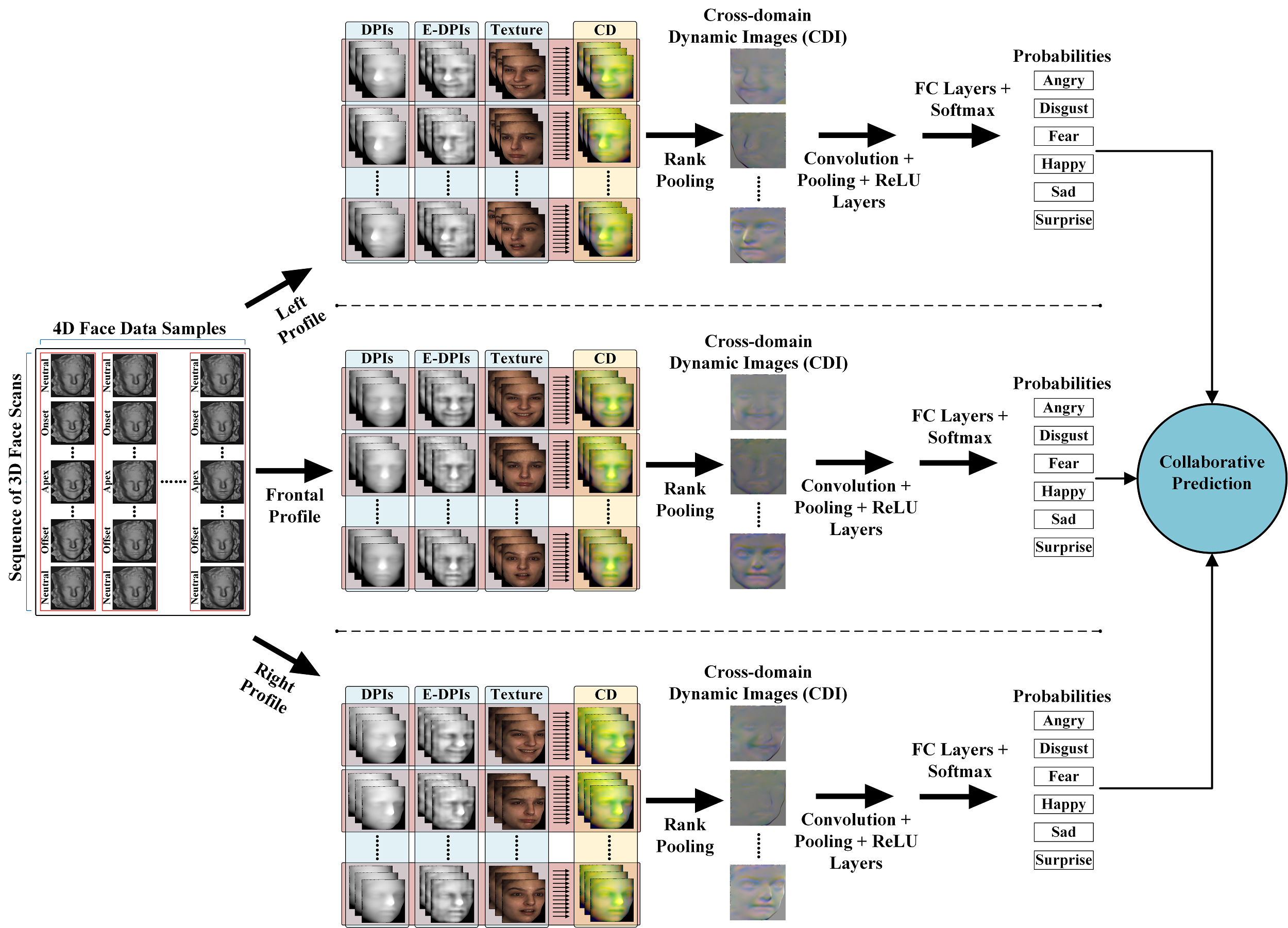}
	\centering
	\caption{The proposed CCDN method for 4D FER.}
	\label{fig:4DFER_floechart}
\end{figure}Consequently, given a 4D dataset with $N$ examples, we process each 3D face scan independently to combat the unavoidable presence of outliers. Therefore, we define
\begin{equation}
\label{eq:1}
I^{4D} = \{I_{nt}^{3D}\},  \text{ } \forall t = \{1,2,3,...,T_n\} \text{ and }\forall n = \{1,2,3,...,N\},
\end{equation}
where $I^{4D}$ represents the set of different 4D data examples, and $I_{nt}^{3D}$ denotes $n$th 3D face video and $t$th temporal scan. Note that (\ref{eq:1})~$\Rightarrow |I^{4D}| = N \text{, and } |I_{nt}^{3D}| = T_n$. For a given 3D shape with $M$ vertices as a $3M \times 1$ vector, we denote its mesh as
\begin{equation}
\label{eq:2}
\textbf{m} = [\textbf{v}_1^T,...,\textbf{v}_M^T]^T = [x_1,y_1,z_1,...,x_M,y_M,z_M]^T,
\end{equation}
where $\textbf{v}_j = [x_j,y_j,z_j]^T$ are the face-centered Cartesian coordinates of the $j$th vertex, and $\textbf{m}_{t}$ is a mesh of each $t$th 3D face frame such that $\textbf{m}_{n} = \{\textbf{m}_{nt}\} \forall n$. Now, since we generate multi-views from each $I_{nt}^{3D}$ for an effective collaboration, we extract various alignment profiles ranging from left to frontal to right. The next step is to crop the regions beyond facial contours with the help of given landmarks. Specifically, we remove everything left to the left-side of the face, right to the right-side of the face, regions below the chin, and regions beyond the nose-tip. To take care of the head-hair, we apply a threshold to trim regions above the forehead. This threshold is a fraction of distance from eyebrows to the tip of nose. For a given 3D face, we represent this as
\begin{equation}
\label{eq:3}
\overline{I_{nt}^{3D}} = \eta_c(\eta_r(I_{nt}^{3D})),
\end{equation}
where $\overline{I_{nt}^{3D}}$ is the cropped and rotated face, while $\eta_c(.)$ and $\eta_r(.)$ refers to the cropping and rotation operation. This is also reflected in the \textit{mesh-space}, and (\ref{eq:2}) updates as follows:
\begin{equation}
\label{eq:4}
\overline{\textbf{m}} = \eta_c(\eta_r(\textbf{m})) = [\textbf{v}_1^T,...,\textbf{v}_{\overline{M}}^T]^T = [x_1,y_1,z_1,...,x_{\overline{M}},y_{\overline{M}},z_{\overline{M}}]^T,
\end{equation}
where $\overline{\textbf{m}}$ is the set of updated vertices such that $\overline{\textbf{m}} \subseteq \textbf{m}$ and $\overline{M} \le M$. The pre-processing in (\ref{eq:3}) and (\ref{eq:4}) take cares of all the outliers and noisy-data that could potentially affect the training process. It also enables the computation of multi-views for every 3D face.

The use of geometrical images have recently increased due to efficient feature mapping from 3D to 2D \cite{7163090}. Therefore, once the raw 3D faces are processed and the outliers removed, we then compute texture images as $f_T: \overline{I^{3D}} \rightarrow I_T$, and depth images (DPI) as $f_D: \overline{I^{3D}} \rightarrow I_D$ from the updated meshes via 3D to 2D rendering, where $f_T$ and $f_D$ represent the function mapping from 3D mesh to texture image $I_T \in \mathbb{R}^{K^2}$ and depth image $I_D \in \mathbb{R}^{K^2}$, respectively, where $K^2$ is the number of pixels. In order to sharpen the images more with richer information about facial deformations, we apply contrast-limited adaptive histogram equalization on DPIs to get enhanced-depth images (E-DPIs) as $I_{ED} = \eta_s(I_D)$, where $\eta_s(.)$ represents the sharpening operator. Consequently, we get pre-processed images of different image domains (\eg, texture and depth) as shown in of Fig. \ref{fig:4DFER_floechart}.

\subsection{Cross-domain Dynamic Images}
\begin{figure}[t!]
	\includegraphics[width=\linewidth]{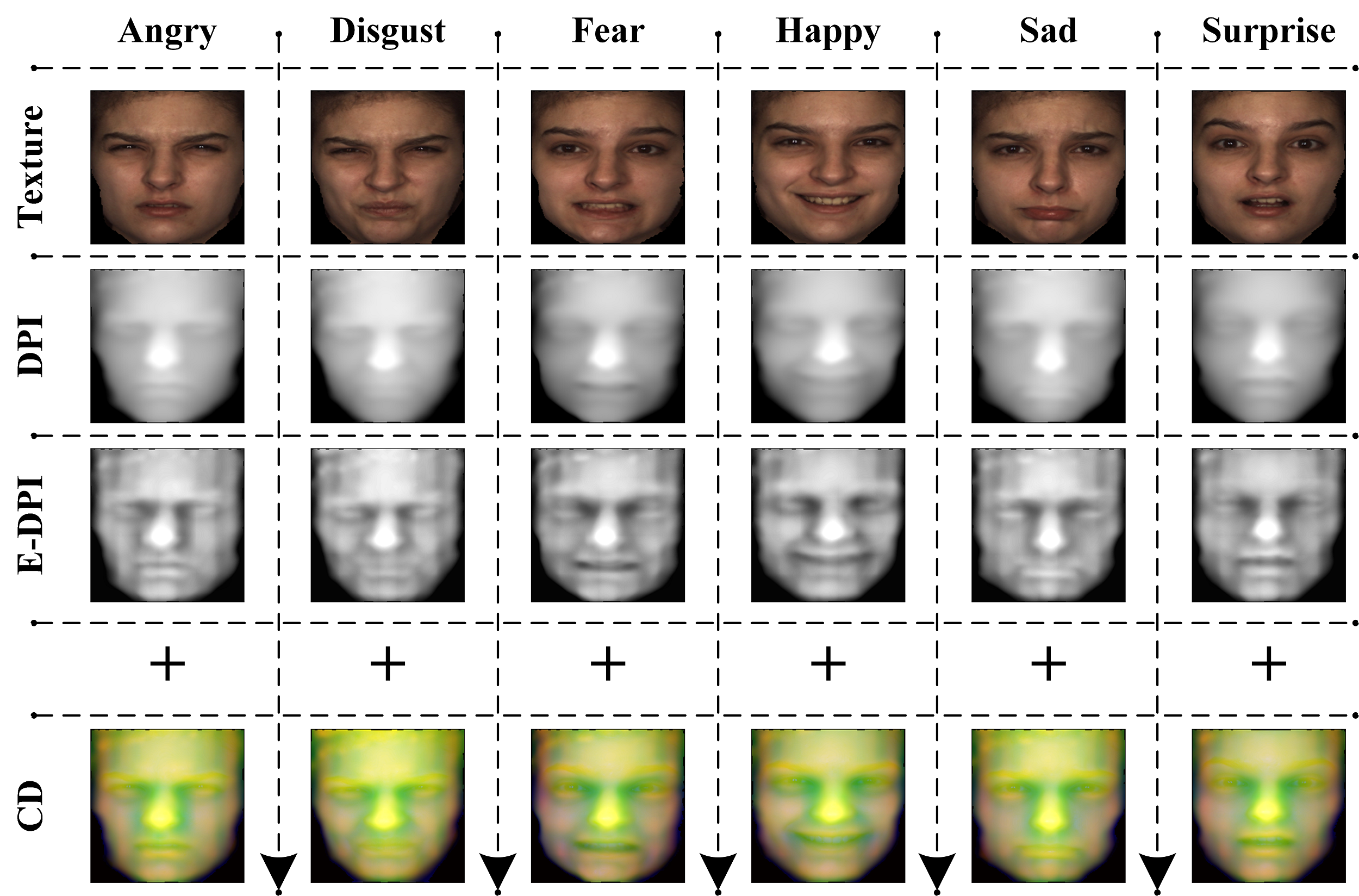}
	\centering
	\caption{Cross-domain images for different prototypical facial expressions.}
	\label{fig:CD_images}
\end{figure}

Instead of independently using the extracted 2D images, we believe that they are correlated, so the deformation patterns from each image domain should be processed collectively. Therefore, we introduce the concept of \textit{cross-domain (CD)} images as shown in Fig. \ref{fig:CD_images}. The CD images are generated by simply adding the contributions of images from different domains into a single image. Consequently, for a given 4D example, we map each 3D frame into its corresponding $I_T, I_D \text{, } I_{ED}$ image domains, and then combine them into cross-domain images $I^{CD} = \{I_{nt}^{CD} \in \mathbb{R}^{K^2} \text{ } \forall n,t\}$. This simple yet effective approach maximize the pattern distance among different classes and minimizes the distance for same class. The effectiveness of the CD images can be supported by the following reasons:
\begin{itemize}
	\item The correlated patterns among different domains are encoded in CD images. This is not applicable when using the extracted images independently from each other. In fact, when not using CD images, the patterns learned from, say DPIs, may negatively influence the prediction scores due to its over-smoothed nature.
	\item The idea of CD images has been never explored and can be a pioneer for not only FER but many other areas including analysis of 4D data for medical diagnosis, action recognition, emotions from gestures, micro-expressions, multi-modal learning, etc.
\end{itemize}

Nevertheless, once the CD image are obtained, we then focus on capturing the spatio-temporal patterns of the resultant CD sequences. For this purpose, performing rank pooling via dynamic images \cite{bilen2018action} serves as an appropriate choice which converts the dynamics of an entire video into a single RGB image. Therefore, we compute the dynamic images for the entire 4D examples using the corresponding sequences of CD images over multi-views. The resultant cross-domain dynamic images (CDIs), as shown in the central part of Fig.~\ref{fig:4DFER_floechart}, store spatio-temporal information by capturing the patterns in a single image which is well-suited for training a deep network. Note that this is not a similar idea as presented by Li \etal \cite{8373807} because they used geometrical images independently, while we use CD images for generating dynamic images and model them collaboratively. CDIs are used in our method to collaborate over multi-views for expression analysis which, to the best of our knowledge, has never been reported in the literature.

\subsection{Collaborative Prediction}
\label{subsec:Collaborative_Prediction}
The aim of the collaborative prediction is to tailor the predictions by considering the voting scores from various resources. Specifically, when the network is trained on input CDIs after multiple Convolution+Pooling+ReLU layers, we use the probability scores of each expression over multi-views in a collaborative fashion. Although the deformation patterns from CDIs significantly distinguish different emotional classes already, the multi-view collaborative approach further improves the probabilities of predicting correct classes. 

For classifying six facial expressions, let us represent the predicted probabilities and collaboratively-updated probabilities, respectively, as
\begin{equation}
\label{eq:6}
C = [\textbf{c}_1^T,...,\textbf{c}_N^T]^T, \text{ and}
\end{equation}
\begin{equation}
\label{eq:7}
C(n,l) = \frac{1}{|\Theta|}\sum_{\forall \theta \in \Theta} C(n,l_\theta), \text{ }\forall n,l.
\end{equation}
Here, $\textbf{c} = \{\rho_l\} \text{ for } l = \{1,...,6\},$ represents the predicted probabilities of the six facial expressions for $n$th example, while $\Theta$ is the set of rotation angles at which the multi-views are captured. The final predictions $F(n)$ are computed as maximum of expression scores
\begin{equation}
\label{eq:8}
F(n) = \text{max}\{C(n)\}, \text{ }\forall n.
\end{equation}

\subsection{4D Augmentation}
\begin{figure}[b!]
	\includegraphics[width=\linewidth]{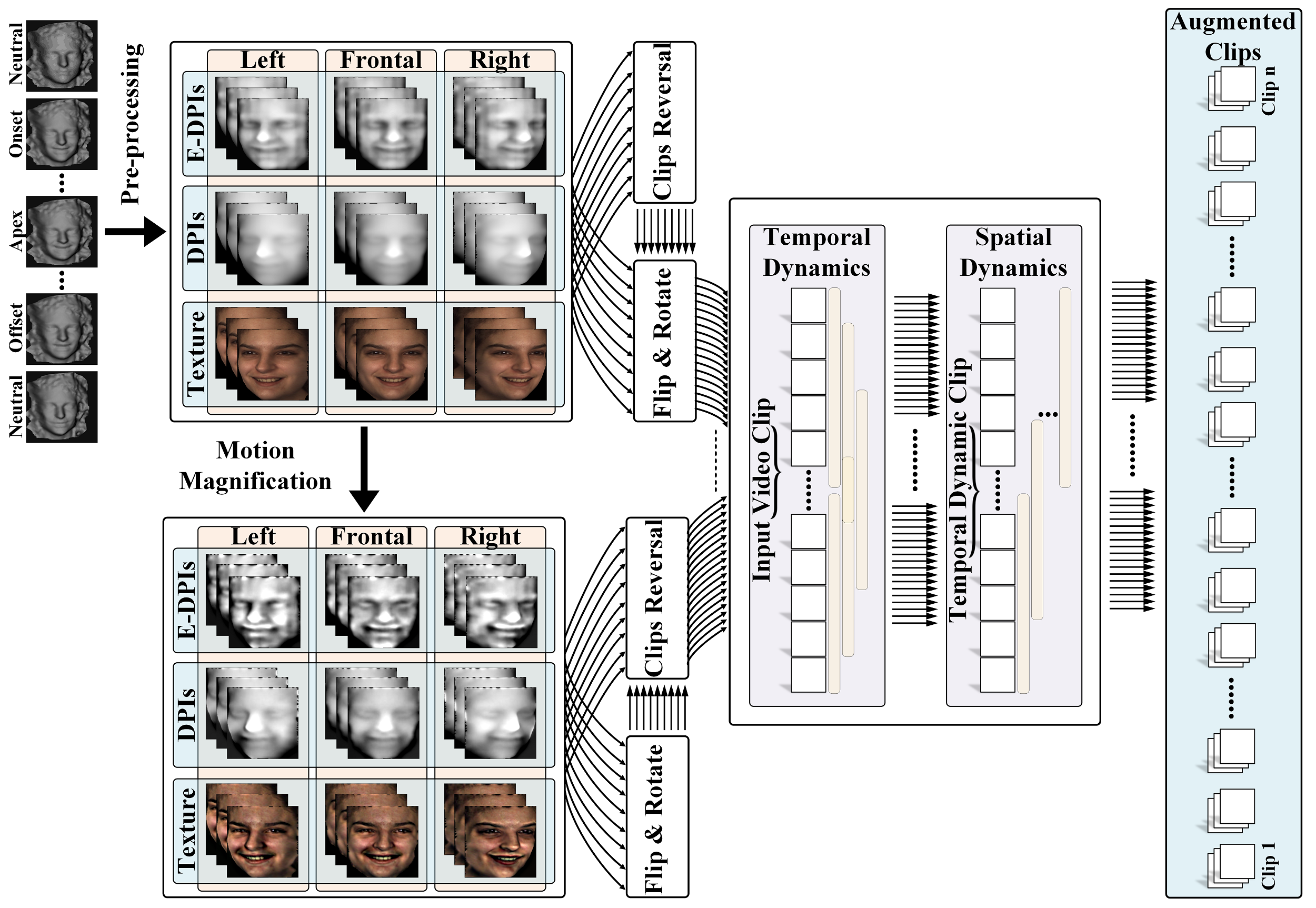}
	\centering
	\caption{Overview of the proposed 4D augmentation method.}
	\label{fig:4DAugmentation}
\end{figure}
Since the training sets for 4D is severely limited, which could potentially restrict the learning performance, there is a need to size up the data for end-to-end training. Some efforts have been made to divide a given video into sub-videos of various strides and window sizes (\eg, see \cite{7780700}). Although this increases the data size for training to some extent, it does not help much for the network in learning different spatio-temporal deformation patterns. It also does not benefit the FC layers that significantly contribute in terms of network weights. Therefore, we propose a novel 4D augmentation method which not only increases the size of training set to a desirable scale but also magnifies the pattern differences of different classes.

As shown in Fig.~\ref{fig:4DAugmentation}, we use five steps to augment our data into a larger set. First, for a reference 4D video, we extract the texture, DPIs and E-DPIs images over multi-views as discussed earlier. Second, we duplicate the extracted set by applying Eulerian Video Magnification (EVM) \cite{Wu12Eulerian} in order to capture the facial muscle movements more precisely. Third, we apply order reversal to both the original and the magnified sets. Fourth, we flip and rotate the facial contents in each frame to get a more enlarged training set. Finally, we apply windowing to extract sub-video clips. Contrary to the existing windowing methods, we first emphasize on temporal dynamics by using a smaller stride and larger window size, and then use a larger stride and smaller window size for capturing spatial dynamics.

\section{Experimental Results and Analysis}
\label{sec:results}
In this section, we analyze our experimental results to evaluate the performance of our proposed 4D FER method. For the experiments, we use the commonly adopted BU-4DFE dataset. It consists of 56 females and 44 males (total 101 subjects) each with six prototypical expressions, i.e., angry, disgust, fear, happy, sad, and surprise, each containing 3D sequences of raw face scans. The frame rate for each 3D video is 25 fps lasting around 3 to 4 seconds.

In order to make fair comparison with previous results, we followed \cite{7780700} and use their experimental setting instead of fine tuning parameters. Specifically, for every expression, we employ a 10-fold subject-independent cross-validation. As well, we report our final 4D FER accuracy results over 10 different iterations each time with different training, validation and testing sets. For our proposed CCDN model, we use the pre-trained VGGNet \cite{Parkhi15} as our CNN model, and extract images from 3D face scans to be size of $224 \times 224$. 

Additionally, we use our CCDN model in the proposed 4D augmentation. Here, we do not use CD representations to independently validate the performance gain of proposed augmentation method. This means that the predicted scores collaborate over multi-views and different image domains to compute updated scores. For EVM, we set the magnification level to the standard value of $\alpha = 4$ \cite{7851001}. All experiments were run on a GP100GL GPU (Tesla P100-PCIE), and the entire training duration takes about half a day.
\begin{table}[b!]
	\begin{center}
		\begin{tabular}{|l|c|}
			\hline
			Method & FER Accuracy ($\%$) \\
			\hline\hline
			2012 Sandbach \etal \cite{sandbach2012recognition} & 64.60 \\ \hline
			2011 Fang \etal \cite{6130440} & 75.82 \\ \hline
			2015 Xue \etal \cite{7045888} & 78.80 \\ \hline
			2010 Sun \etal \cite{Sun:2010:TVF:1820799.1820803} & 83.70 \\ \hline
			2016 Zhen \etal \cite{7457243} & 87.06 \\ \hline
			2018 Yao \etal \cite{Yao:2018:TGS:3190503.3131345} & 87.61 \\ \hline
			2012 Fang \etal \cite{FANG2012738} & 91.00 \\ \hline
			2018 Li \etal \cite{8373807} & 92.22 \\ \hline
			2014 Ben Amor \etal \cite{amor20144} & 93.21 \\ \hline
			2016 Zhen \etal \cite{8023848} & 94.18 \\ \hline
			\textbf{Ours (without CD)} & \textbf{84.70}\\ \hline
			\textbf{Ours} & \textbf{96.50}\\ \hline
		\end{tabular}
	\end{center}
	\caption{Accuracy comparison with the state-of-the-art on the BU-4DFE dataset.}
	\label{table:4DFERresults}
\end{table}

We compare the FER accuracies of our method and several state-of-the art methods \cite{sandbach2012recognition,6130440,7045888,Sun:2010:TVF:1820799.1820803,7457243,Yao:2018:TGS:3190503.3131345,FANG2012738,8373807,amor20144,8023848} on the BU-4DFE dataset in Table \ref{table:4DFERresults}. Results show that our method outperforms the existing ones in terms of the expression classification accuracy. Note that without using CD in the CCDN model and by collaborating over all predicted scores, the accuracy reaches only $84.7\%$ due to incorrect utilization of the correlated information across different image domains. For instance, due to over-smoothed DPIs, its dynamic images can not encode temporal deformations efficiently therefore misguides the predictions. In contrast, an accuracy of $96.5\%$ is achieved with our proposed method using CD.

\begin{figure}[t!]
	\includegraphics[width=0.7\linewidth]{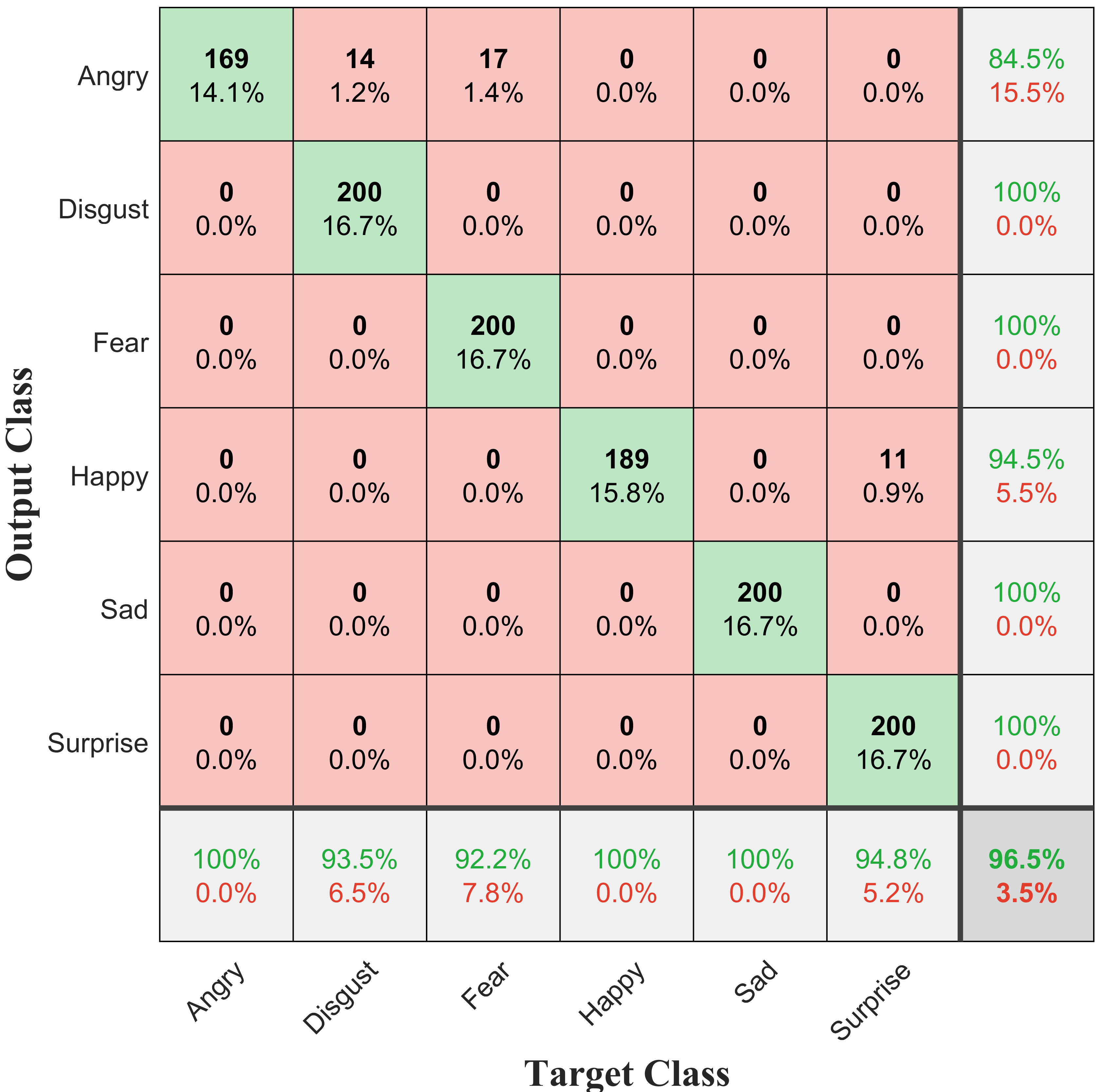}
	\centering
	\caption{Confusion matrix of recognizing six prototypic expressions on BU-4DFE database.}
	\label{fig:4DFER_CM}
\end{figure}

Furthermore, we also show the confusion matrix of our experiment in Fig. \ref{fig:4DFER_CM}. As shown, our method predicts the correct emotions in most cases. Error cases are mainly among fear, angry and disgust emotions which are also reported in the literature \cite{7045888} as easily confused. Nevertheless, our method still predicts the angry emotion correctly in $84.5\%$ of the time.

Finally, to validate the effectiveness of our 4D augmentation method, we compare it when augmentation was not used at all or partially used. As shown in table \ref{table:augresults}, the FER accuracy increases from $84.7\%$ to $87.6\%$ when the model is trained on  magnified clips and its variants (rotated, flipped, and windowed), and the resultant prediction scores are obtained collaboratively over multi-views and different image domains. However, note the $3.3\%$ drop in this accuracy because the original clips were not used. This is because the magnified videos also contain noisy observations which influence the prediction performance if used without the original clips. On the other hand, it can be seen that $93.8\%$ accuracy is achieved when the full augmentation method is used, thanks to the collaborative nature of our method.

\section{Conclusions}
\label{sec:conclusions}
\begin{table}[b!]
	\begin{center}
		\begin{tabular}{|l|c|}
			\hline
			Augmentation Levels (without CD) & FER Accuracy ($\%$) \\
			\hline\hline
			Original & 84.70 \\ \hline
			EVM + Variants & 87.60 \\ \hline
			Original + Variants & 90.90 \\ \hline
			\textbf{All (Original + EVM + Variants)} & \textbf{93.80}\\ \hline
		\end{tabular}
	\end{center}
	\caption{FER Accuracy ($\%$) comparison of 4D augmentation on the BU-4DFE dataset.}
	\label{table:augresults}
\end{table}
In this paper, we proposed an automatic method for 4D FER using Collaborative Cross-domain Dynamic Image Network (CCDN). Correlated patterns from the extracted geometrical images were used to capture facial movements. Additionally, we generated cross-domain dynamic images that encode the temporal dynamics in a single image via rank pooling. The collaboration performed over multi-views is an added benefit of our FER method. Moreover, we also introduced a novel 4D augmentation method that expands the training set as well as introduces more patterns for improved FER. Our method achieved an accuracy of $96.5\%$ outperforming the state-of-the-art 4D FER methods on the commonly used BU-4DFE dataset under widely adopted experimental settings proving its effectiveness.

\bibliographystyle{unsrt}
\bibliography{BMVC_arXiv}

\end{document}